\documentclass[lettersize,journal]{IEEEtran}
\usepackage{amsmath,amsfonts}
\usepackage{algorithmic}
\usepackage{algorithm}
\usepackage{array}
\usepackage[caption=false,font=normalsize,labelfont=sf,textfont=sf]{subfig}
\usepackage{textcomp}
\usepackage{stfloats}
\usepackage{url}
\usepackage{verbatim}
\usepackage{graphicx}
\usepackage{hyperref}
\usepackage{cite}
\usepackage{booktabs}
\begin{document}

\title{OSDMamba: Enhancing Oil Spill Detection from Remote Sensing Images Using Selective State Space Model}

\author{Shuaiyu Chen,
        Fu Wang,
        Peng Ren,
        Chunbo Luo,
        Zeyu Fu

\thanks{This work was supported by the China Scholarship Council and University of Exeter PhD Scholarships.}
\thanks{Shuaiyu Chen, Fu Wang, Chunbo Luo and Zeyu Fu are with the Department of Computer Science, University of Exeter, Exeter, UK. Emails: sc1321@exeter.ac.uk, fw377@exeter.ac.uk, c.luo@exeter.uk, z.fu@exeter.ac.uk (\textit{Corresponding author: Zeyu Fu})}
\thanks{Peng Ren is with the College of Oceanography and Space Informatics, China University of Petroleum (East China). Email: pengren@upc.edu.cn}
}



\maketitle

\begin{abstract}
Semantic segmentation is commonly used for Oil Spill Detection (OSD) in remote sensing images. However, the limited availability of labelled oil spill samples and class imbalance present significant challenges that can reduce detection accuracy. Furthermore, most existing methods, which rely on convolutional neural networks (CNNs), struggle to detect small oil spill areas due to their limited receptive fields and inability to effectively capture global contextual information.
This study explores the potential of State-Space Models (SSMs), particularly Mamba, to overcome these limitations, building on their recent success in vision applications. We propose OSDMamba, the first Mamba-based architecture specifically designed for oil spill detection. 
OSDMamba leverages Mamba’s selective scanning mechanism to effectively expand the model’s receptive field while preserving critical details. Moreover, we designed an asymmetric decoder incorporating ConvSSM and deep supervision to strengthen multi-scale feature fusion, thereby enhancing the model's sensitivity to minority class samples. Experimental results show that the proposed OSDMamba achieves state-of-the-art performance, yielding improvements of 8.9\% and 11.8\% in OSD across two publicly available datasets. The source codes will be made publicly available at  \url{https://github.com/Chenshuaiyu1120/Oil-Spill-detection}.


\end{abstract}


\section{Introduction}
Marine oil spills, primarily caused by drilling platform blowouts~\cite{1}, pipeline leaks~\cite{2}, and tanker spills~\cite{3}, pose a severe threat to coastal and marine ecosystems. 
Effectively mitigating the environmental impact of oil spills requires the quick identification of spill locations\cite{solberg2007oil}, which remains a challenging task.
Taking advantage of the rapid advancements in optical imaging sensors and synthetic aperture radar (SAR),  providing a large amount of earth observation data, using deep learning-based models has become a promising solution for automatically detecting the oil spill~\cite{HTSM,brekke2005oil,mados}.
Convolutional Neural Networks (CNN) based approaches \cite{satyanarayana2023oil,ronneberger2015u,li2023ds} have shown promising performance in oil spill detection, outperforming traditional threshold and machine learning methods~\cite{fiscella2000oil,guo2014oil,temitope2020advances,li2023ds,dehghani2023oil}. Their encoder-decoder architectures effectively capture high-level semantic features, enhancing detection accuracy. 
Transformers ~\cite{transformer1,transformer2} utilise self-attention to model relationships among all image patches simultaneously, enabling the direct capture of global contextual information for efficient object segmentation. However, their high computational complexity poses challenges for oil spill detection, especially in real-world marine applications. 
Recently, Wu~\cite{wu2024compositional} proposed a compositional oil spill detection framework, SAM-OIL, which adapted SAM with an ordered mask fusion module, achieving higher segmentation accuracy.

\begin{figure}[t]
  \centering
  \centerline{\includegraphics[width=9cm]{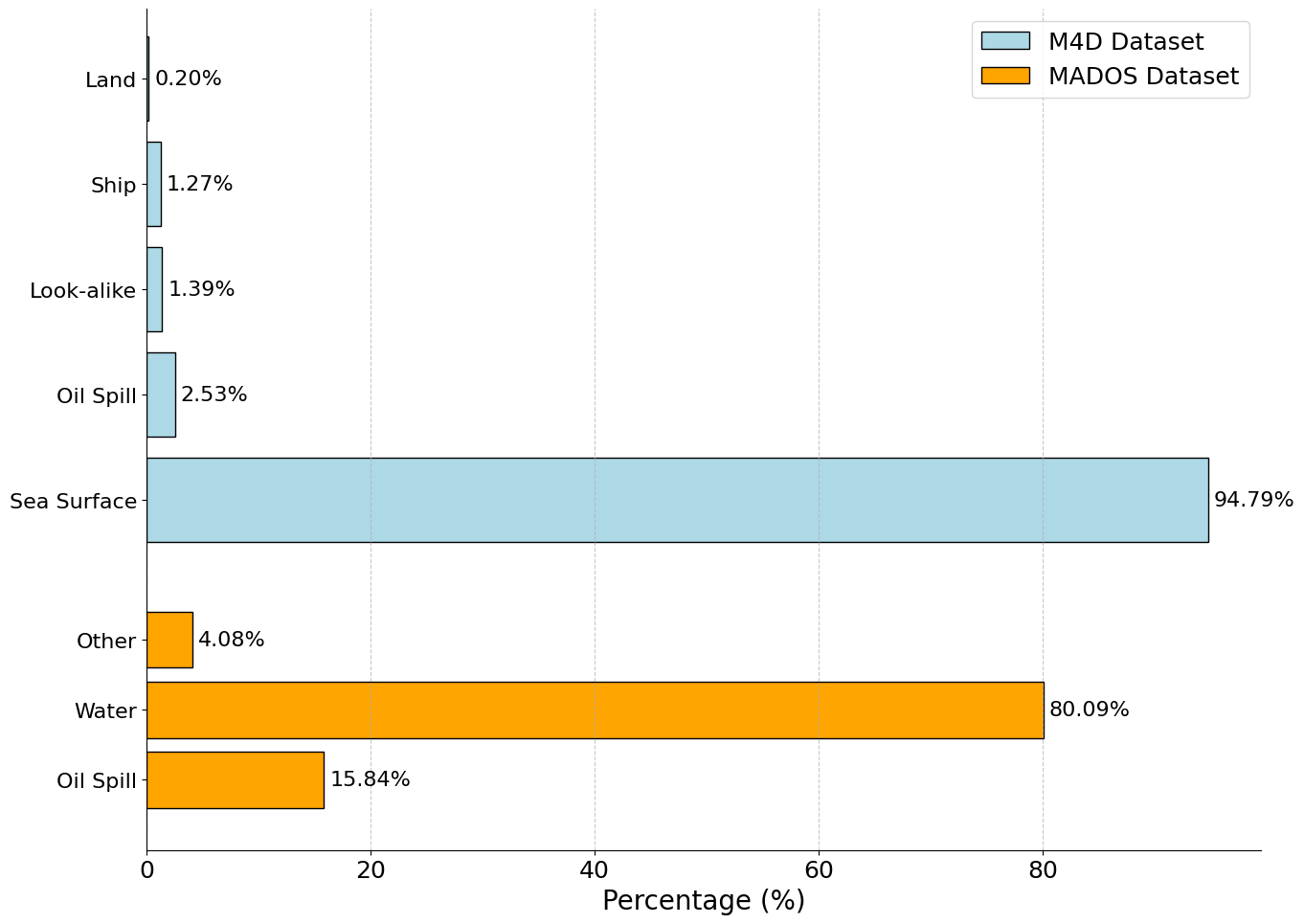}}
 \vspace{-1.5em}
  \caption{The distribution of the semantic classes in the Oil spill detection dataset~\cite{M4D} and MADOS Dataset~\cite{mados}. (
For ease of presentation, we merged the 20 categories of the MADOS dataset into "Oil Spill," "Water," and "Other.")}\label{fig:data distribution}
\end{figure}


\begin{figure*}[t]
  \centering
    \centerline{\includegraphics[width=18cm]{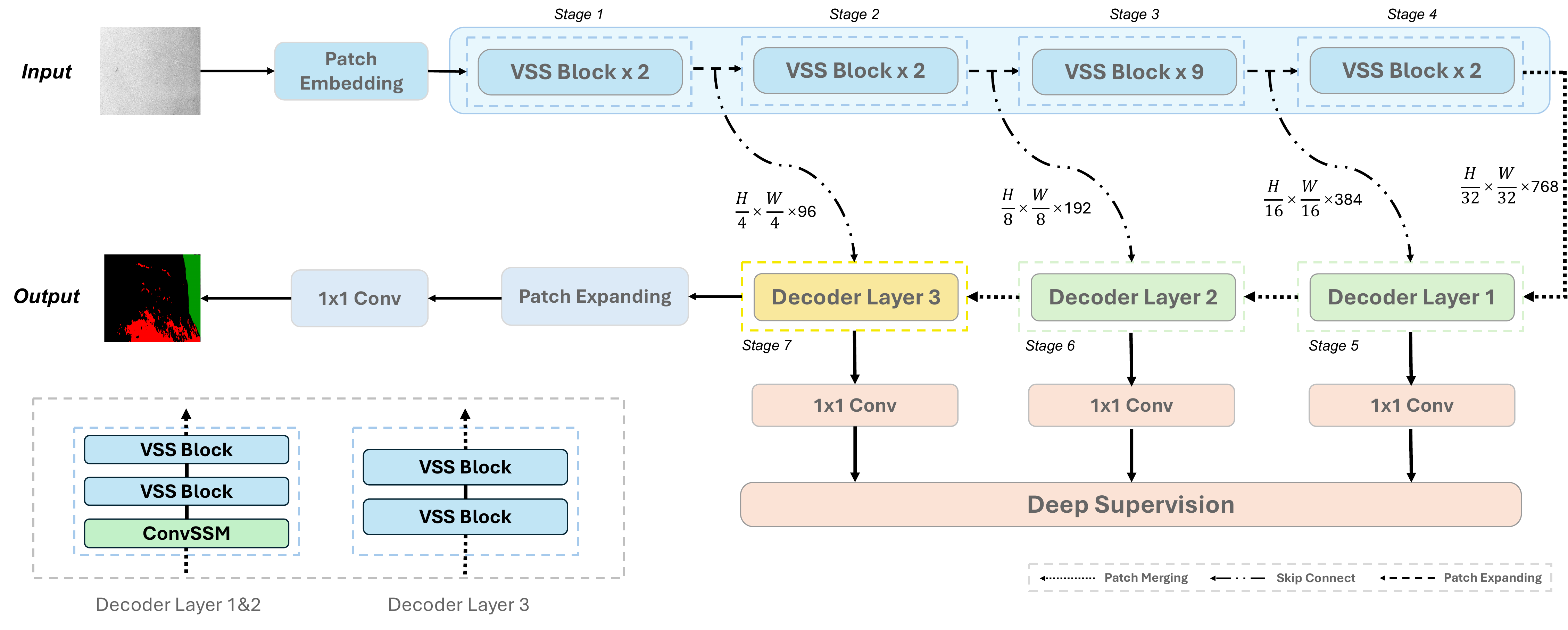}}
 \vspace{-1.0em}
   \caption{Architectural overview of the proposed method. OSDMamba uses a pre-trained, modified VMamba as the encoder. The decoder consists of two parts: one made up of ConvSSM-VSS Block-VSS Block, and the other composed of two VSS Blocks.}\label{fig:mambabet}
\end{figure*}

\begin{figure}[t]
  \centering
  \centerline{\includegraphics[width=8cm]{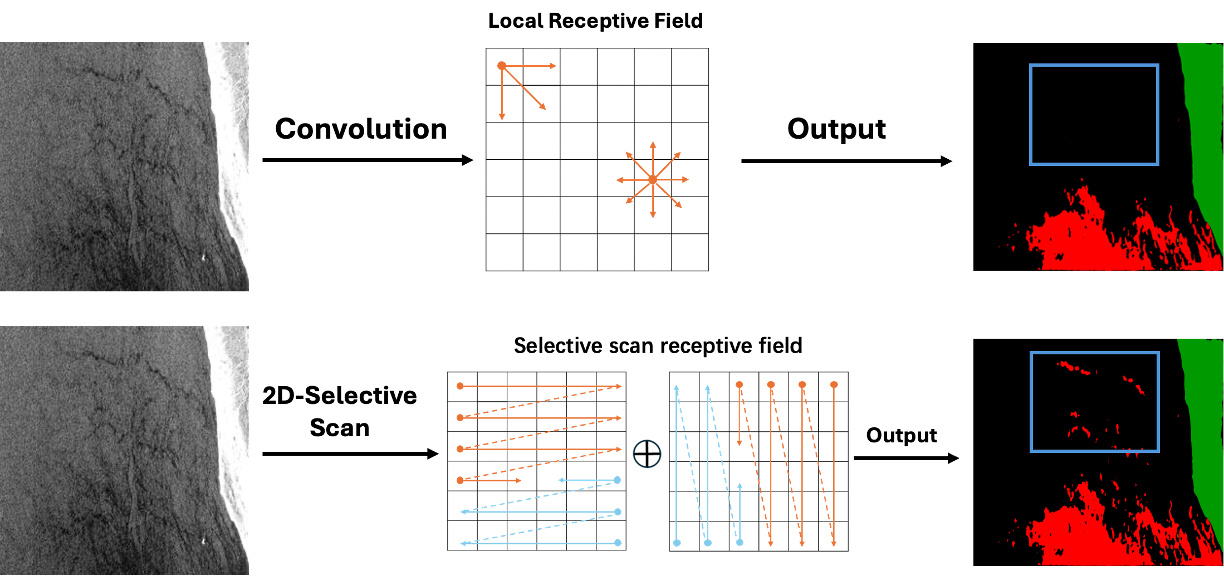}}
  \caption{Comparison of receptive fields between convolution and 2D selective scan used in OSDMamba.
  (The blue bounding boxes indicate that some small objects with convolution  are misclassified, while the selective scan  effectively addresses this issue.)}
  \label{fig:fp}
\end{figure}

Despite the recent success, there remain two key challenges that affect detection performance: class imbalance and limited receptive fields of convolution.  Oil spill incidents are rare, resulting in a limited number of available image samples \cite{brekke2005oil}. In these images, oil spills occupy only a small portion of the scene, leading to a significant class imbalance between the sea surface and oil spill regions, as shown in Fig \ref{fig:data distribution}. Additionally, convolution-based models have limited receptive fields. When the receptive field is small, the model may fail to capture global information and the contextual relationships of the target, leading to misclassification or complete omission of small objects, as shown in Fig.~\ref{fig:fp}.




Recently, state-space models (SSMs), such as the Mamba deep learning model ~\cite{gu2023mamba}, have shown promise in addressing various imaging tasks \cite{liu2024swin,ruan2024vm}. Mamba excels at capturing long-range dependencies and extracting meaningful features at both local and global scales, leveraging bidirectional scanning and large convolutional kernels for enhanced performance. In this paper, we study the effectiveness of Mamba models to the above challenges of oil spill detection. To achieve this, we propose the first Mamba-based architecture tailored for detecting oil spills, called \textbf{OSDMamba}, as shown in Fig. \ref{fig:mambabet}. 
By leveraging Mamba's sliding window mechanism for local self-attention, it ensures the dependence of information in each local region, helping the model more accurately learn minority class features in imbalanced datasets \cite{zhu2024vision}. Additionally, we develop an asymmetric decoder with ConvSSM \cite{smith2024convolutional}
and deep supervision for more effective feature fusion at multiple scales. By combining convolution operations with SSM, the model preserves global state information while extracting local features, enhancing feature representations in complex scenarios. 
Experimental results show that the proposed OSDMamba achieves state-of-the-art performance, yielding improvements of 8.9\% and 11.8\% in oil spill detection across two publicly available datasets.
\section{Methodology}
Fig. \ref{fig:mambabet} illustrates our proposed OSDMamba. A SAR image is first divided into small patches using a Patch Embedding technique. These patches are transformed into a linear embedding sequence, which is further processed through Visual State Space (VSS) Blocks with a configuration of \{2, 2, 9, 2\}~\cite{zhu2024vision}. Features with downsampling rates of \{4, 8, 16\}, generated at the corresponding three stages, are transferred to the decoder for upsampling. We employ an asymmetric decoder structure, where ConvSSM modules are applied in the first two stages to achieve spatial feature fusion. 


\subsection{Encoder}
\subsubsection{Vision State Space Block}

Building on insights from~\cite{liu2024swin,ruan2024vm}, we utilize the VSS block as the foundational unit of OSDMamba. The VSS Block~\cite{zhu2024vision}, a core component of the Mamba network, addresses long-range dependency modelling by leveraging local and global information through a state-space model (SSM). 
To handle 2D image data, the VSS Block utilizes a 2D Selective Scan (SS2D) method, as shown in Fig. \ref{fig:fp}, unfolding image patches in four distinct directions to create separate sequences. These sequences are then processed by the SSM, and the resulting features are merged to reconstruct a comprehensive 2D feature map.
Given an input feature map \( \mathbf{Z} \in \mathbb{R}^{H \times W \times C} \), the SS2D output is defined as:
\begin{equation}
    \mathbf{Z}_v = \mathcal{S}_6(\text{Expand}(\mathbf{Z}, v)), \quad v \in \{1, 2, 3, 4\},
\end{equation}
\begin{equation}
    \mathbf{Z} = \text{Merge}(\mathbf{Z}_1, \mathbf{Z}_2, \mathbf{Z}_3, \mathbf{Z}_4),
\end{equation}
where \( v \) represents the scanning direction, \( \text{Expand}(\cdot) \) and \( \text{Merge}(\cdot) \) denote the scan expansion and merging operations~\cite{liu2024swin}, and \( \mathcal{S}_6 \) is the core SSM operator in the VSS Block.



\subsubsection{Encoder Structure} 

We based the OSDMamba encoder on design principles from VMamba-Tiny\cite{zhu2024vision} and Swin-UMamba\cite{liu2024swin}. Similar to Swin-UMamba, OSDMamba performs 2× downsampling at each stage to preserve low-level details.
The later stages follow a similar design of VMamba-Tiny, with 2, 2, 9, and 2 VSS blocks in stages 1 to 4, respectively. This configuration balances model depth and efficiency, allowing for robust feature extraction across multiple scales.
Additionally, both the VSS blocks and patch merging layers can be initialized with ImageNet~\cite{imagenet} pre-trained weights to enhance performance and accelerate convergence.

\subsection{Decoder}
\subsubsection{Convolutional State Space Model}
ConvSSM integrates convolutional operations with state space dynamics to model spatiotemporal sequences efficiently. The state is updated by:
\begin{equation}
\mathbf{X}_k = \mathbf{A} * \mathbf{X}_{k-1} + \mathbf{B} * \mathbf{U}_k,\quad 
\mathbf{Y}_k = \mathbf{C} * \mathbf{X}_k + \mathbf{D} * \mathbf{U}_k
\end{equation}
To allow parallel scans across time, the state kernel $\mathbf{A}$ is restricted to pointwise convolution:
\begin{equation}
\mathbf{A} \in \mathbb{R}^{P \times P \times 1 \times 1}
\end{equation}
This design yields sub-quadratic complexity $\mathcal{O}(UPk^2HWL)$ while maintaining spatial structure. Furthermore, ConvSSM is mathematically equivalent to a block-diagonal SSM, enabling HiPPO-based initialization for long-range temporal modeling. Our variant, ConvS5, leverages these properties to support fast and accurate sequence modeling at scale. ConvSSM bridges the gap between local convolutional priors and global sequence modeling, enabling efficient prediction in high-resolution temporal tasks.
\begin{table*}[t]\small
\small
\setlength{\tabcolsep}{3pt}
\centering\caption{The quantitative performance comparison and ablation experiments of the proposed method (on M4D Dataset).\label{tab:table1}}
\centering
\begin{tabular}{c c c c c c c}
\toprule
Model & SeaSurface(\%, \(\uparrow\)) & Oil Spill(\%, \(\uparrow\)) & Look-alike(\%, \(\uparrow\)) & Ship(\%, \(\uparrow\)) & Land(\%, \(\uparrow\)) & mIoU(\%, \(\uparrow\))\\
\midrule

Unet~\cite{ronneberger2015u} & 93.90 & 53.79 & 39.55 & 44.93 & 92.68 & 64.97 \\

LinkNet~\cite{Linknet} & 94.99 & 51.53 & 43.24 & 40.23 & 93.97 & 64.79 \\

PSPNet~\cite{PSPNet} & 92.78 & 40.10 & 33.79 & 24.42 & 86.90 & 55.60 \\

Deeplabv2~\cite{Deeplabv2} & 94.09 & 25.57 & 40.30 & 11.41 & 74.99 & 49.27 \\

Deeplabv2(msc)~\cite{Deeplabv2} & 95.39 & 49.28 & 31.26 & 88.65 & 93.97 & 62.83 \\

Deeplabv3+~\cite{dpv3p} & 96.43 & 53.38 & 55.40 & 27.63 & 92.44 & 65.06 \\

YOLOv8-SAM ~\cite{wu2024compositional} & 94.34 & 41.84 & 48.15 & 52.48 & 87.65 & 64.89 \\

SAM-OIL  ~\cite{wu2024compositional} & 96.05 & 51.60 & \textbf{55.60} & \textbf{52.55} & 91.81 & 69.52 \\



\midrule
OSDMamba without Decoder &95.60 & 64.76&46.30&50.32&92.07& 67.60\\ 
OSDMamba without VSS Block &92.80 & 64.53& 46.58& 50.11& 91.80& 67.28\\ 
OSDMamba without ConvSSM &93.11 & 64.94& 46.14& 50.62& 92.24& 67.83\\ 

OSDMamba without Deep Supervision &95.50& 65.97&53.61&43.20&91.93 & 69.01\\
OSDMamba (ours) & \textbf{96.47} & \textbf{65.59} & 47.57 & 46.85 & \textbf{94.76} & \textbf{70.25} \\
\bottomrule
\end{tabular}
\end{table*}


\subsubsection{Decoder Structure}

Although existing decoders~\cite{liu2021swin} can preserve fine details and global context, their performance remains limited under highly imbalanced class distributions. To address this, we propose a locally asymmetric decoder with two customized upsampling blocks for different decoding stages. In early stages, where local details are crucial for small oil spill detection~\cite{fiscella2000oil}, we introduce the VSS Block and ConvSSM: the former maintains spatial boundaries and minority features, while the latter enhances global context and local extraction through self-attention and convolution. For later stages, we adopt a lightweight design with patch expansion and dual VSS blocks. This tailored upsampling strategy improves detection in imbalanced and small-target scenarios. To further strengthen multi-level representations, we apply deep supervision~\cite{liu2024swin} using $1 \times 1$ convolutions on decoder outputs at 1/4, 1/8, and 1/16 scales.

\section{Experiments}
\subsection{Datasets}
We use the M4D dataset~\cite{krestenitis2019oil} and the MADOS dataset~\cite{mados} for evaluation. 
The M4D dataset consists of 1,002 training SAR images and 110 testing images.
The images in the M4D dataset have a resolution of 1250 × 650 pixels and include five semantic categories: sea surface, oil spill, oil spill look-alike, ship, and land. 
The MADOS is a globally distributed benchmark designed for detecting oil spills and debris.
This dataset contains 174 high-resolution multispectral Sentinel-2 satellite images collected between the year 2015 and 2022, with approximately 1.5 million labelled pixels spanning 15 thematic categories. 

\subsection{Implementation details}

The proposed OSDMamba model was implemented using PyTorch on an NVIDIA L40 GPU, following the training strategy outlined in \cite{HTSM}. We utilized the AdamW optimizer\cite{adamw} with an initial learning rate of 0.01 and a weight decay of 0.0001.  We trained the OSDMamba model using a hybrid loss function, as given by
$L = -\alpha_t (1 - p_t)^\gamma \log(p_t) + (1 - \frac{|A \cap B|}{|A \cup B|})$
where $\alpha_t$ is the class weight, used to balance class imbalance; $P_t$ is the predicted probability of the true class by the model; $\gamma$ is a modulation factor used to adjust the weight of hard and easy samples; $A$ is the predicted positive region; $B$ is the ground truth positive region. The model was initialized with pre-trained weights from ImageNet\cite{imagenet}. A batch size of 4 was used, and the model was trained for 100 epochs across all stages.

\subsection{Comparison with State-of-the-Art}
Tab.~\ref{tab:table1} presents a quantitative comparison of OSDMamba against other state-of-the-art methods on the M4D dataset. 
Overall, OSDMamba achieves a mIoU of 70.25\%, outperforming all other models in the comparison.  This result highlights OSDMamba’s superior ability to detect marine oil spills with enhanced generalization performance.
Notably, OSDMamba delivers the best performance in the oil spill category, achieving a 12.18\% improvement over the second-best model, U-Net. This significant gain indicates OSDMamba’s capability to effectively learn from underrepresented classes in imbalanced datasets. Furthermore, OSDMamba also achieves top performance in detecting Sea Surface and Land, demonstrating its robustness across multiple categories. 

Tab.~\ref{tab:mados} provides detailed results on the MADOS dataset. Here, OSDMamba again surpasses all competing methods, with an improvement of 0.6\% in F1-score and 8.9\% in mIoU. These consistent improvements across different datasets confirm OSDMamba’s ability to deliver reliable and accurate detection performance in various scenarios. At the same time, as shown in Tab. III, our model outperforms transformer-based counterparts of similar scale in terms of both parameter count and computational cost, achieving higher accuracy while maintaining superior efficiency.



\begin{table}[t]
\centering
\caption{Quantitative comparison of OSDMamba on the MADOS dataset}
\begin{tabular}{lccc}
\toprule
model & F1(\%, \(\uparrow\)) & mIoU(\%, \(\uparrow\)) & OA(\%, \(\uparrow\)) \\
\midrule
RF   & 56.6 &	43.9 & 67.1 \\
RF++\cite{RF++}   & 64.4  & 52.4 & 83.8 \\
U-Net\cite{ronneberger2015u}   & 63.8  & 51.0 & 82.9 \\
SegNext\cite{segnext}   & 60.6  & 49.2 & \textbf{86.6} \\
MariNeXt (reproduced)\cite{mados} & 70.6  & 59.2 & 81.6 \\
OSDMamba (ours)   & \textbf{71.2}  & \textbf{68.1} & 82.3 \\
\bottomrule
\end{tabular}
\label{tab:mados}
\end{table}


\begin{table}[h]
\centering
\caption{Comparison of model complexity and false positives.}
{
\begin{tabular}{lcc}
\toprule
\multicolumn{3}{c}{\textbf{Model size and FLOPs}} \\
\midrule
\textbf{Model} & \textbf{Params (M)} & \textbf{FLOPs (G)} \\
Mix Transformer~\cite{wang2021mixedtransformerunetmedical} & 286 & $\approx$389 \\
OSDMamba        & 110 & $\approx$270 \\
\midrule
\multicolumn{3}{c}{\textbf{False positives on MADOS dataset}} \\
\midrule
\textbf{Model} & \textbf{Oil FP(\%,↓)} & \textbf{Overall FP(\%,↓)} \\
Baseline (U-Net) & 37.2 & 43.5 \\
OSDMamba         & \textbf{32.4} & \textbf{35.8} \\
\bottomrule
\end{tabular}
}
\label{tab:combined}
\end{table}

\subsection{Ablation Study}
We conducted a comprehensive ablation study to quantify the individual contributions of key components within OSDMamba, as presented in Tab.~\ref{tab:table1}. The removal of the decoder leads to a noticeable decline in performance (mIoU: 67.60\%), highlighting the significance of the proposed multi-scale feature fusion. Eliminating the VSS Block further degrades performance (mIoU: 67.28\%), demonstrating the importance of long-range dependency modeling for effectively capturing contextual information in sparse target regions. The exclusion of ConvSSM slightly reduces performance (mIoU: 67.83\%), indicating that although state-space modeling via convolution is beneficial, its impact is partially complemented by the VSS Blocks. The absence of deep supervision results in a substantial drop in the segmentation of minority categories such as Ship (53.61\% to 43.20\%), and a decrease in overall performance (mIoU: 69.01\%), affirming its role in guiding multi-scale representation learning. By contrast, the complete OSDMamba configuration achieves the highest accuracy (mIoU: 70.25\%), outperforming all ablated variants and confirming the effectiveness of our architectural design in enhancing segmentation robustness, particularly under class-imbalanced conditions.

To highlight how our method addresses the two previously mentioned challenges, we present the false positive (FP) statistics both within the oil spill areas and overall, in Tab. III.
Compared to the baseline method, our approach significantly reduces the number of false positives (FP) within minority classes, particularly in oil spill and land regions. As shown in Tab. III, our Mamba-based model achieves consistently higher IoU and recall scores for these underrepresented categories, indicating enhanced robustness to class imbalance.
This improvement benefits from the structural advantages of Mamba, which leverages long-range state space modeling to aggregate sparse signals across the spatial domain, in contrast to CNNs that rely on local receptive fields. Additionally, the state space mechanism naturally preserves broader contextual information and facilitates smoother token interactions, stabilizing the representation of rare classes.

\begin{figure}[t]
  
    \centerline{\includegraphics[width=8.9cm]{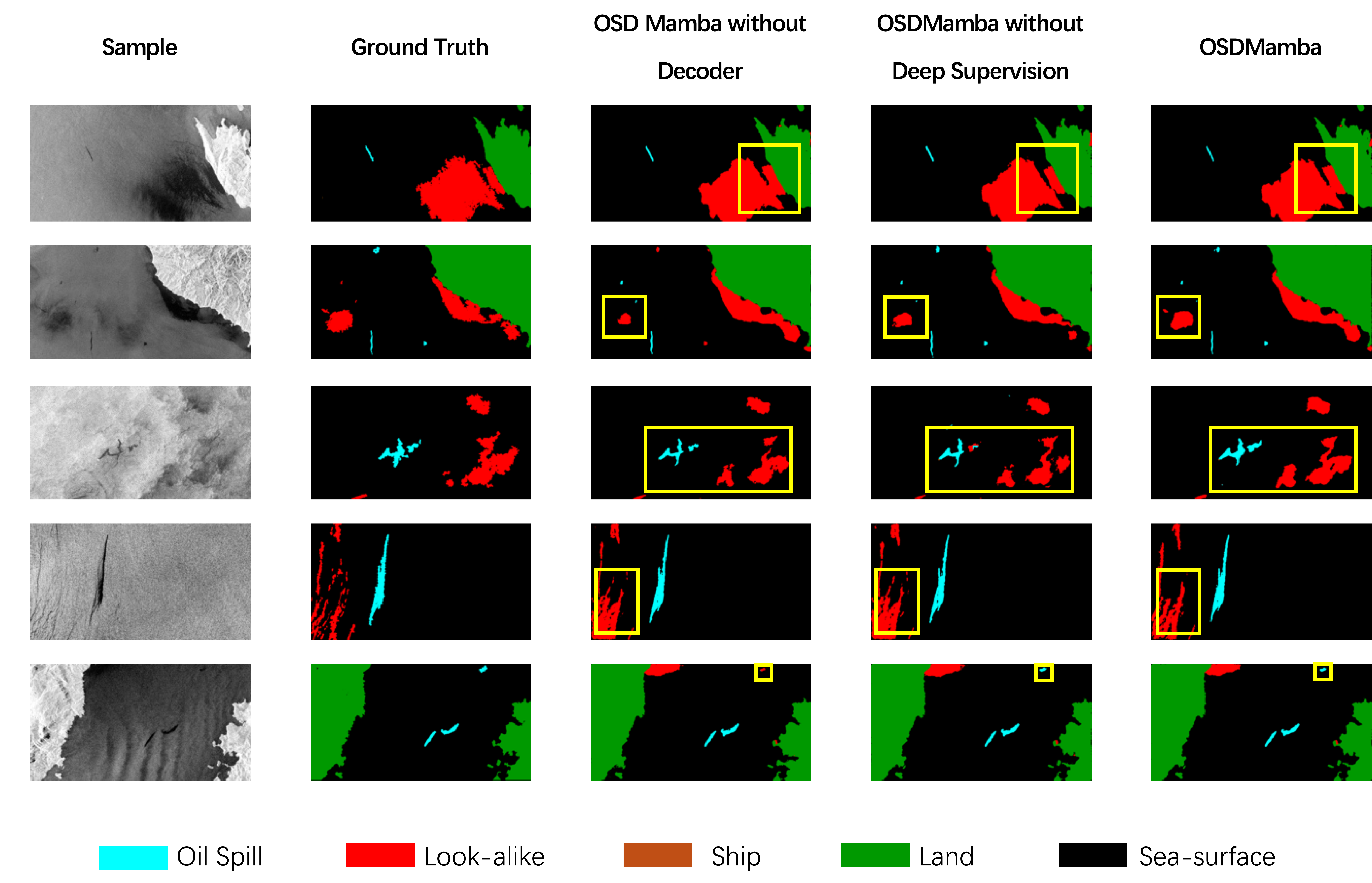}}
\centering
\caption{Qualitative analysis of ablation experiments}
\label{fig:Result visualization}
\end{figure}
\subsection{Qualitative Analysis}

As shown in Fig. \ref{fig:Result visualization}, the qualitative analysis visualises the contributions of each module through a column-wise comparison. Using the CNN decoder, the model struggles with segmenting small or ambiguous regions, such as isolated oil spills and look-alike areas (e.g., third columns).
Our decoder improves the segmentation of minority classes, particularly for oil spills and ships (e.g., 5th column), but some regions remain under-segmented or misclassified.
The complete OSDMamba achieves the best segmentation, accurately capturing fine details and distinguishing small-scale targets, as depicted in the 5th column of Fig. \ref{fig:Result visualization}.

\section{Conclusion}
In this paper, we presented a new Mamba-based image segmentation framework, OSDMamba, for marine oil spill detection. By stacking VSS Blocks, the model is better able to distinguish limited samples. Additionally, we designed an asymmetric decoder and integrated ConvSSM into specific decoder layers to enhance feature fusion. Extensive experiments demonstrate that our model achieves state-of-the-art segmentation performance on two oil spill detection datasets. We believe that our design using the Mamba architecture offers a new perspective for marine oil spill detection tasks.

\end{document}